\documentclass[letterpaper, 10 pt, conference]{ieeeconf}  

\IEEEoverridecommandlockouts                              

\usepackage{graphicx}
\usepackage{epsfig}
\usepackage{amsmath}
\usepackage{amssymb}
\usepackage[ruled]{algorithm2e}
\usepackage{multirow}
\usepackage{array}
\usepackage{ctable} 
\usepackage{makecell}
\usepackage[capposition=bottom]{floatrow}
\usepackage{comment}
\DeclareMathAlphabet{\mathcal}{OMS}{cmsy}{m}{n}
\usepackage{textcomp}
\usepackage{dblfloatfix}
\usepackage{caption}
\usepackage{subcaption}
\usepackage{colortbl}
\usepackage{booktabs}
\usepackage{mathtools}
\usepackage{textpos}
\usepackage{wrapfig,lipsum,booktabs}
\usepackage[normalem]{ulem}
\usepackage{titlesec}
\usepackage{hyperref}
\titlespacing*{\section}{8pt}{8pt plus 2pt minus 2pt}{4pt plus 1pt minus 1pt}
\titlespacing*{\subsection}{4pt}{4pt plus 1pt minus 1pt}{2pt plus 0pt minus 0pt}

\usepackage{siunitx}

\newcommand{\name}{\textsc{Sonicverse}\xspace}

\definecolor{customgray}{rgb}{0.9, 0.9, 0.9}
\newcolumntype{g}{>{\columncolor{customgray}}c}
\newcolumntype{z}{>{\columncolor{customgray}}l}
\newcolumntype{?}[1]{!{\vrule width #1}}
\renewcommand{\paragraph}[1]{{\vspace{1mm}\noindent\textbf{#1}\,\,}}

\title{\name: A Multisensory Simulation Platform \\ for Embodied Household Agents that See and Hear }

\author{Ruohan Gao$^{*}$, Hao Li$^{*}$, Gokul Dharan, Zhuzhu Wang, Chengshu Li, Fei Xia \\Silvio Savarese, Li Fei-Fei, Jiajun Wu \\
Stanford University
\thanks{\vspace{-0.2in}*Equal contribution, in alphabetical order}
}

\begin{document}

\maketitle
\thispagestyle{empty}
\pagestyle{empty}

\begin{abstract}
Developing embodied agents in simulation has been a key research topic in recent years. Exciting new tasks, algorithms, and benchmarks have been developed in various simulators. However, most of them assume deaf agents in silent environments, while we humans perceive the world with multiple senses. We introduce \name, a multisensory simulation platform with integrated audio-visual simulation for training household agents that can both see and hear. \name models realistic continuous audio rendering in 3D environments in real-time. Together with a new audio-visual VR interface that allows humans to interact with agents with audio, \name enables a series of embodied AI tasks that need audio-visual perception. For semantic audio-visual navigation in particular, we also propose a new multi-task learning model that achieves state-of-the-art performance. In addition, we demonstrate \name's realism via sim-to-real transfer, which has not been achieved by other simulators: an agent trained in \name can successfully perform audio-visual navigation in real-world environments. Sonicverse is available at: \url{https://github.com/StanfordVL/Sonicverse}.
\end{abstract}
\section{Introduction}
\label{sec:intro}

Future household robots should be able to see, hear, and follow voice instructions from humans. They should find their way around your home, go where you need them, and fetch the desired object at your command. For example, a robot should be able to locate the owner by its voice and follow the speaker, go to the kitchen to attend to an accident when hearing a cracking sound, and deliver a bottle of water when hearing your request from the bedroom.

An increasing amount of work in embodied AI has thus studied visual navigation, where an agent must use its egocentric visual stream to intelligently move around, explore a new space that has not been mapped before, and navigate to its goal. The goal can be specified by a point~\cite{gupta2017cognitive,habitat19iccv,wijmans2019dd,Chaplot2020Learning}, an object~\cite{zhu-iccv2017,chaplot_object_2020,savinov2018semi}, or a room~\cite{zhou2019towards}. Many state-of-the-art simulators~\cite{habitat19iccv,Xiang_2020_SAPIEN,shen2020igibson,robothor,puig2018virtualhome,gan2020threedworld,zhu2020robosuite,james2020rlbench,chen2020soundspaces,li2021igibson} are also designed for developing embodied AI agents to tackle various challenging tasks.

Despite the encouraging progress, there is a salient missing ingredient---the environment is silent, and the agents cannot hear. Limited prior work has tackled embodied learning with audio~\cite{chen2020soundspaces,gan2019look,chen2021waypoints,chen2021semantic,purushwalkam2020audio,gao2020visualechoes}: SoundSpaces~\cite{chen2020soundspaces} and ThreeDWorld~\cite{gan2020threedworld} are two notable exceptions. However, SoundSpaces uses a separate audio dataset of pre-computed room impulse responses at a discrete grid of spatial locations with a pre-defined height, which prevents sampling data at new locations; ThreeDWorld supports continuous-space audio rendering, but it assumes a box-shaped approximation for modeling 3D environments, which limits its realism.

\begin{figure*}[ht]
    \center
    \includegraphics[scale=0.49]{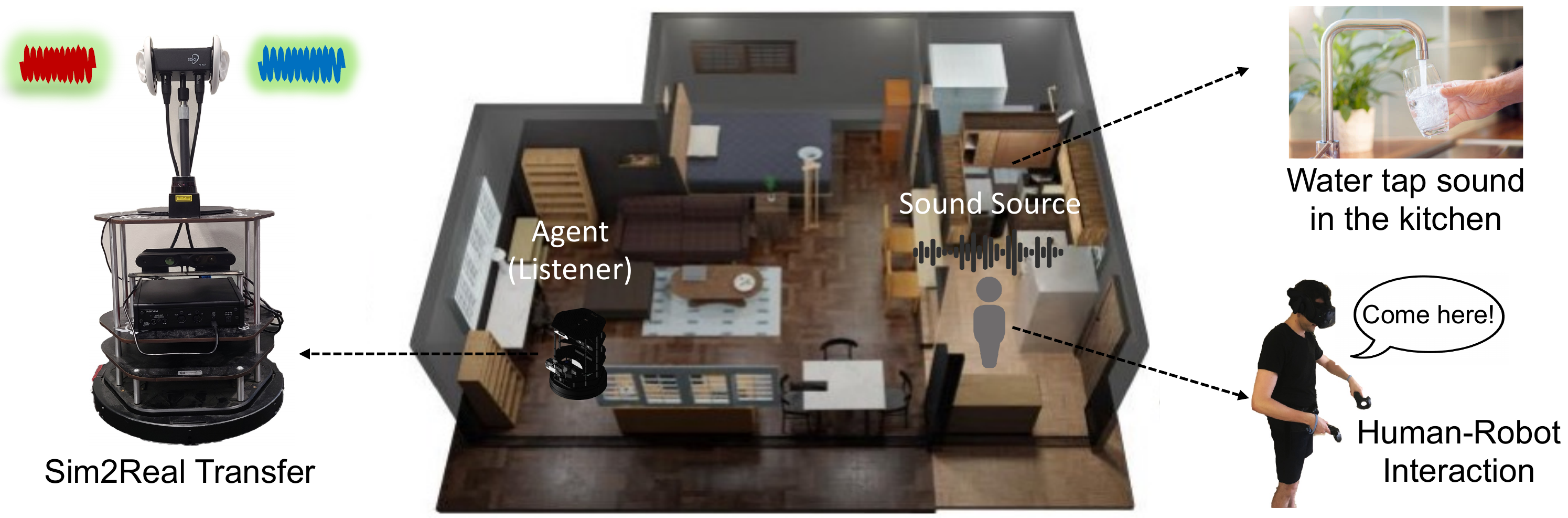}
    \caption{Illustration of our \name simulation platform. We visualize the top-down map of a 3D environment we support for spatial audio rendering. The source audio can either be a sound coming from a semantically meaningful object or room (e.g., water tap sound in the kitchen), or a voice from a person wearing the VR headset. The agent in the environment can act as a listener to receive directional information about the sound source and perform tasks that require audio-visual perception. An audio-visual embodied agent trained in \name's environments can also generalize to their real-world counterparts (a TurtleBot with a binaural microphone shown on the left). 
    }
    \label{fig:simulator_summary}
    \vspace{-0.05in}
\end{figure*}

We introduce \name, a new multisensory simulation platform for training audio-visual embodied agents that overcome these limitations. Not only can \name render audio over a continuous space in real-time, but it also achieves high fidelity spatial audio rendering by using the complete scene geometry and material properties. As shown in Figure~\ref{fig:simulator_summary}, we can attach semantically meaningful sounds to existing object assets or rooms in a 3D environment (e.g., water tap sound in the kitchen), and have agents serve as listeners who can receive the transmitted audio in space. Furthermore, we also support audio streaming with a Virtual Reality (VR) interface, enabling many potential applications for voice-driven human-robot interaction. For example, we can instantiate a person wearing the VR headset 
as an audio-visual avatar in the simulated environment. The person can issue a voice command, and the robot can locate its position from the spatial cues of the binaural audio it receives, and navigate to the person as instructed.

As a case study, we tackle the semantic audio-visual navigation task with continuous audio, and use our \name simulation platform as a testbed for training navigation agents. We propose a multi-task learning framework for both audio-visual navigation and occupancy map prediction. The goal of occupancy map prediction is to infer occupancy (free or occupied) for unseen regions based on the audio-visual context. Our key insight is that these two tasks are mutually beneficial: the better the audio-visual state feature learned for navigation, the more useful it is for occupancy prediction; the prediction of occupancy in turn regularizes the learning for audio-visual navigation. We also train a model for audio-goal navigation and successfully deploy agents trained in simulation to real-world environments.

Our contributions are threefold. First, we introduce \name, a new multisensory simulation platform that models continuous audio rendering in 3D environments in real-time, providing a new testbed for many embodied AI and human-robot interaction tasks that need audio-visual perception. Second, we introduce a multi-task learning framework for semantic audio-visual navigation and occupancy map prediction, which achieves state-of-the-art results. Third, we are the first to show that audio-visual navigation agents trained in simulation can be successfully deployed in real-world environments.

\section{Related Work}\label{sec:related}

\paragraph{Embodied AI Simulators.} A series of well-designed simulation environments and benchmarks have been introduced for embodied AI research in the past several years: Habitat~\cite{habitat19iccv,szot2021habitat}, iGibson~\cite{shen2020igibson,li2021igibson}, AI2Thor~\cite{kolve2017ai2}, RoboTHOR~\cite{robothor}, Sapien~\cite{Xiang_2020_SAPIEN}, Robosuite~\cite{zhu2020robosuite}, VirtualHome~\cite{puig2018virtualhome}, RLBench~\cite{james2020rlbench}, Meta-World~\cite{yu2020meta}, etc. These simulators above have enabled many new tasks and possibilities for training and developing embodied agents. However, none of them support multisensory perception.

The simulators that are most related to ours are SoundSpaces~\cite{chen2020soundspaces} and ThreeDWorld~\cite{gan2020threedworld}, each with distinct features and limitations. SoundSpaces can only render sounds at a discrete grid of spatial locations where the pre-computed room impulse responses are available; ThreeDWorld assumes a user-specified box-shaped approximation of rooms for rendering spatial audio, which limits its realism. Our \name simulator can both render continuous audio in 3D spaces in real-time, and achieve high realism by using the complete scene geometry and surface material properties. Concurrent with our work, SoundSpaces 2.0~\cite{chen22soundspaces2} also supports on-the-fly continuous audio rendering. While their platform focuses more on supporting visual-acoustic learning and simulation results, we are the first to show that agents trained in a simulator can be successfully deployed in real-world environments for audio-visual navigation.

\paragraph{Visual Navigation.} Significant progress has been made in the realm of visual navigation in recent years. Given the visual sensory data from onboard sensors, an embodied agent is tasked to reach a point-goal~\cite{gupta2017cognitive,habitat19iccv,wijmans2019dd,Chaplot2020Learning}, object-goal~\cite{zhu-iccv2017,chaplot_object_2020,savinov2018semi,batra2020objectnav}, or room-goal~\cite{zhou2019towards}, or to follow language instructions~\cite{zhao2021evaluation,anderson2018vision,chen2021topological}. Some methods~\cite{wijmans2019dd,hochreiter1997long} perform end-to-end training with implicit memory structure to predict actions directly from pixels, whereas others leverage semantic priors~\cite{chaplot_object_2020} and explicit memory structure like metric or topological maps~\cite{chen2021topological} to facilitate navigation. Most existing visual navigation research is conducted without audio, an important gap that \name aims to fill in.

\paragraph{Embodied Audio-Visual Learning.} Recent work uses both vision and audio for a variety of embodied AI tasks, such as audio-visual navigation~\cite{chen2020soundspaces,gan2019look,chen2021waypoints,chen2021semantic}, floor plan reconstruction~\cite{purushwalkam2020audio}, representation learning~\cite{gao2020visualechoes}, curiosity-driven exploration~\cite{dean2020see,gan2020noisy}, or object-centric learning~\cite{gao2021ObjectFolder,gao2022ObjectFolderV2}. Our work offers a new testbed to support these embodied AI tasks that require audio-visual perception, and we also perform a case study on the audio-visual navigation task to demonstrate the usefulness and realism of our simulator.

\paragraph{Audio-Visual Learning from Videos.} Videos inherently contain both sights and sounds. Recent inspiring work leverage both modalities for a variety of interesting tasks, including self-supervised representation learning~\cite{owens2016ambient,arandjelovic2017look,owens2018audio,Korbar2018cotraining}, audio-visual source separation~\cite{gao2018objectSounds,zhao2018sound,gan2020music,gao2019coseparation,tzinis2020into}, sound source localization in images~\cite{arandjelovic2017look,Senocak_2018_CVPR,tian2018audio, hu2020discriminative}, and spatial audio generation~\cite{gao2019visualsound,morgadoNIPS18,garg2021geometry}. Unlike any of them, our work aims to enable embodied AI tasks that require audio-visual perception.
\section{The \name Simulation Platform}\label{sec:simulator}

We introduce \name, a new embodied AI simulation platform that supports audio-visual perception. \name is built on top of iGibson 2.0~\cite{li2021igibson}, which is an open-source interactive simulation environment for fast visual and physics simulation with a focus on household tasks. We augment it with the powerful open-source spatial audio SDK\footnote{\url{https://github.com/resonance-audio}} of Resonance Audio~\cite{gorzel2019efficient} to enable audio-visual perception of the agents. Below, we introduce the main components of audio simulation, the 3D environments, and the key features of our \name simulation platform.

\begin{figure*}
    \center
    \includegraphics[width=0.95\textwidth]{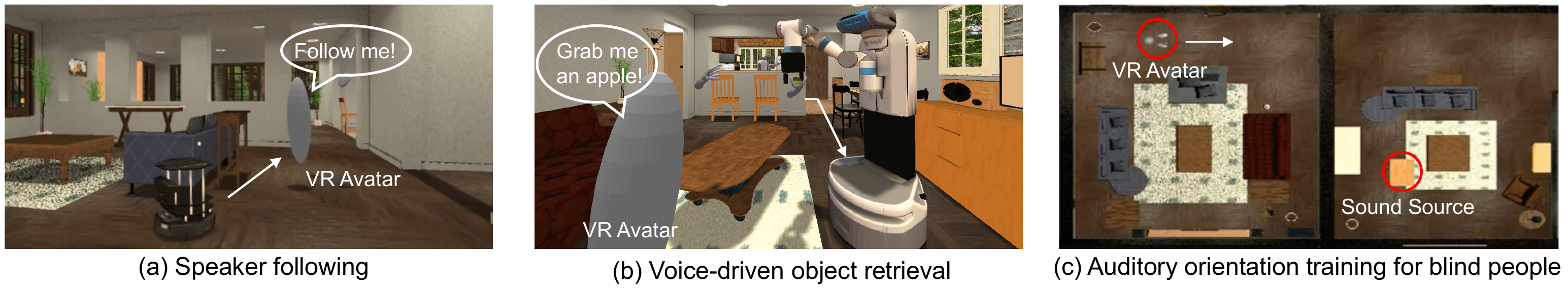}
    \caption{Illustrations of task prototypes enabled by our audio-visual virtual reality interface. See Supp. video for demos.
    \vspace{-0.1in}}
    \vspace{-0.1in}
    \label{fig:VR_examples}
\end{figure*}

\subsection{Acoustic Simulation}
We begin with the main components of audio simulation in~\name. Please see Supp. video for a demonstration.
 
\paragraph{Direct Sound.}
Direct sound represents the sound that travels from the source to the listener without being obstructed or reflected by the environment. This component is attenuated as the distance between source and listener increases, as the energy of direct sound falls exponentially with distance. We also render near-field effects to get the most physically correct amplitude boost (including low-frequency boost) of a near-field source.

\paragraph{Dynamic Occlusion.}
Physical obstructions between the source and the listener attenuate the sound rendered to the listener by means of an occlusion node in Resonance Audio, which blocks high frequencies more than low frequencies to mirror real-world occlusion effects. At each simulation step, we calculate the number of objects that lie on a ray between the source and the listener, and attenuate the received audio accordingly.

\paragraph{Early Reflections and Late Reverberations.}
We initialize a user-specified number of reverb probes in the scene. These probes are uniformly spread throughout the scene, and we run a pre-simulation reverb-baking process by raycasting from each probe and measuring the characteristics of the early and late ray reflections. These properties vary depending on the size and shape of the immediate area as well as the material properties of the surfaces in the scene. Since these properties are pre-computed for each probe and are not updated dynamically, we perform the reverb-baking process on a version of the mesh that contains only the static structures of the scene. During simulation, \name uses the reverb probe that is closest to the listener and renders reverberation effects by taking advantage of the pre-computed rt60s. Early reflections are rendered on-the-fly by also taking into account the listener's position relative to the probe, using a box-shaped approximation of the room. This different treatment of simulating early reflections and late reverberations allows both realistic geometry-based spatial audio rendering and real-time performance.

\paragraph{Head-Related Transfer Functions (HRTFs).}
Humans physically locate sound sources by taking advantage of time and level differences between the sound perceived by each ear. We utilize the HRTFs that incorporate these effects for rendering realistic binaural audio for the listeners. 

\subsection{3D Environments}
Though built on top of iGibson 2.0, \name is flexible and supports two datasets of 3D scenes: Matterport3D~\cite{Matterport3D} and iGibson~\cite{shen2020igibson}. We discuss their characteristics and how we configure them for audio-visual simulation below.

\textbf{Matterport3D~\cite{Matterport3D}:} We use 85 large Matterport3D scenes of real-world homes and other indoor environments with 517$m^2$ of floor space on average. Since Matterport3D scenes are static, we use the entire scene when performing reverb-baking. To do so, we map the semantic mesh categories to Resonance Audio material types (e.g., ``wall'' maps to ``concrete block, painted'', ``curtain'' maps to ``curtain, heavy''), which determine the acoustic properties of the room surfaces such as scattering and absorption coefficients. This enables realistic reverberation and early-reflection modeling when raycasting from points in the scene.

\textbf{iGibson~\cite{shen2020igibson}:} We use 15 fully interactive scenes of real-world homes with furniture and articulated objects. As objects in iGibson scenes are movable, we only use the static skeleton of the scene to perform reverb-baking. This involves mapping the walls and ceiling to ``concrete block, painted'', assigning windows to ``glass'', and floors to ``wood panel''.  

\subsection{Key Features}
Next, we introduce the two key features of \name: the audio-visual Virtual Reality (VR) interface and the capability of Sim2Real transfer. Then we highlight our main differences against the two state-of-the-art audio-visual simulators.

\paragraph{Audio-Visual Virtual Reality Interface.} We augment the VR interface from iGibson 2.0~\cite{li2021igibson} with streaming audio support, which is compatible with major commercially available VR headsets through OpenVR~\cite{OpenVR}. It enables the embodiment of a person wearing the VR headset as an audio-visual avatar and allows an agent to hear human-voiced commands in VR, opening many possibilities for human-robot interaction tasks with audio-visual perception. In Fig.~\ref{fig:VR_examples}, we illustrate three task prototypes enabled by our audio-visual VR interface: 1) \emph{Speaker following}, where the agent needs to follow a person's voice and move together with the target person; 2) \emph{Voice-driven object retrieval}, where the agent needs to receive the voice command from a person, sense its location from the binaural audio cues, fetch the right object as instructed, and finally deliver the object to the person who issues the command; 3) \emph{Auditory orientation training for blind people}, where a blind person wears the VR headset and practices auditory orientation perception by navigating to the sound source in virtual environments. See Supp. video for demos of these applications.

\begin{figure}
    \center
    \includegraphics[width=\textwidth]{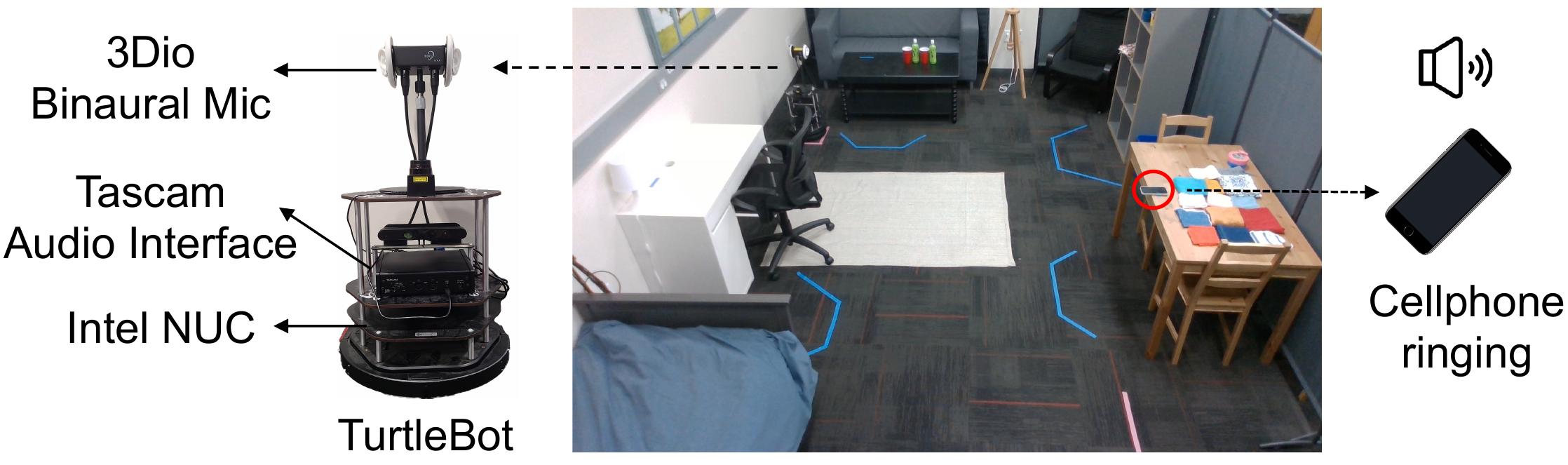}
    \caption{Sim2Real set-up of a TurtleBot equipped with a binaural microphone in a real-world environment.
    \vspace{-0.2in}}
    \vspace{-0.1in}
    \label{fig:Sim2Real}
\end{figure}

\paragraph{Sim2Real Transfer.}
We use a TurtleBot as the embodied agent in our simulator, and we set up its real-world counterpart as shown in Fig.~\ref{fig:Sim2Real}. We mount a 3Dio FS binaural microphone on the real TurtleBot and use a Tascam audio interface to process the received audio. The TurtleBot is equipped with an Asus XTION PRO RGBD camera, and an onboard Intel NUC. This allows us to verify that our audio simulation is realistic, such that the policies learned in \name can be transferred to real-world environments.

\paragraph{Comparing to State-of-the-Art Audio-Visual Simulators.} In comparison to SoundSpaces~\cite{chen2020soundspaces}, a key advantage of our simulator is that it integrates audio and visual simulation by having sounds attached to dynamic objects in the scene. For example, given a loudspeaker playing music in the scene, an agent moving the speaker would hear the audio source move concurrently. Moreover, our interactive scenes allow for dynamic occlusion, making the sound intensity increase or decrease in response to the opening or closing of a door, for example. Our simulator also renders audio over a continuous space, whereas SoundSpaces uses a grid of discrete rendering points throughout the scene. ThreeDWorld~\cite{gan2020threedworld} also uses a built-in version of Resonance Audio in UNITY, though with a user-specified box-shaped approximation. Our implementation achieves higher realism by using the complete scene geometry and automatically mapped materials for reverb-baking. We also demonstrate sufficient audio rendering fidelity for successful Sim2Real transfer. ThreeDWorld does not support spatial audio rendering for VR, though it does support basic sound rendering without advanced spatialization. We do not directly model object impact sounds as in ThreeDWorld, but \name supports the integration of existing multisensory object assets with pre-computed audio simulation~\cite{gao2021ObjectFolder,gao2022ObjectFolderV2}. 

\section{Training Audio-Visual Embodied Navigation Agents in~\name}
\name supports many embodied AI tasks that require audio-visual perception. We tackle the challenging \emph{semantic audio-visual navigation}~\cite{chen2021semantic} task as a case study to demonstrate the usefulness of our simulator. This is a more challenging version of audio-goal navigation~\cite{chen2020soundspaces,gan2019look}, in which an agent must locate a consistently-sounding source. In semantic audio-visual navigation, objects make sounds consistent with their real-world counterparts (e.g., doors make creaking sounds), and these sounds only last for a short period of time. The agent must therefore be able to localize the sound source well after it has stopped emitting sound, perhaps by leveraging the learned knowledge about which objects can emit certain sounds.

\paragraph{Task Definition.} In this task, an agent is required to navigate to a specific semantically meaningful object in an unseen and unmapped environment by hearing the sound emitted by that object. The sound can be non-periodic, discontinuous, and of varied length. To reach the target object, the agent has to reason about the semantic category of the sounding object as well as the binaural spatial cues from audio perception. We use a TurtleBot as the agent for our experiments. We use 15 semantically meaningful sounds used in~\cite{chen2021semantic}, including the sounds from the sink, cushion, tv, shower, etc. Each sound is one-to-one mapped to a specific target category. To be considered successful, the agent needs to locate the target position even after the sound stops, and navigate to the specific target object that was making the sound instead of any objects within the category. 

\paragraph{Action and Observation Spaces.} In contrast to the task's existing specification~\cite{chen2021semantic}, which uses a discrete set of fixed-step translations and rotations, we use a continuous action space over robot wheel velocities. This makes the task setting more realistic and challenging, and more applicable to real-world robotics settings. The agent's observations include an RGB image, a depth map, the binaural audio received at its two ears, a bump sensor input, and its current pose with respect to the starting location.

\paragraph{Episode Specification and Success Criterion.} Each episode is defined by the following: the scene, the start position and orientation of the agent, the target category, a target object within the category, and eight positions sampled within one meter of the target object position, which are considered as the nearby locations that define the object boundary. The agent is considered to meet the success criterion when it reaches any of the nine terminal positions: the eight positions near the target and the original target object position. The distance tolerance for reaching the terminals is 0.36m, which is the width of the real TurtleBot.

\begin{figure*}
    \center
    \includegraphics[scale=0.5]{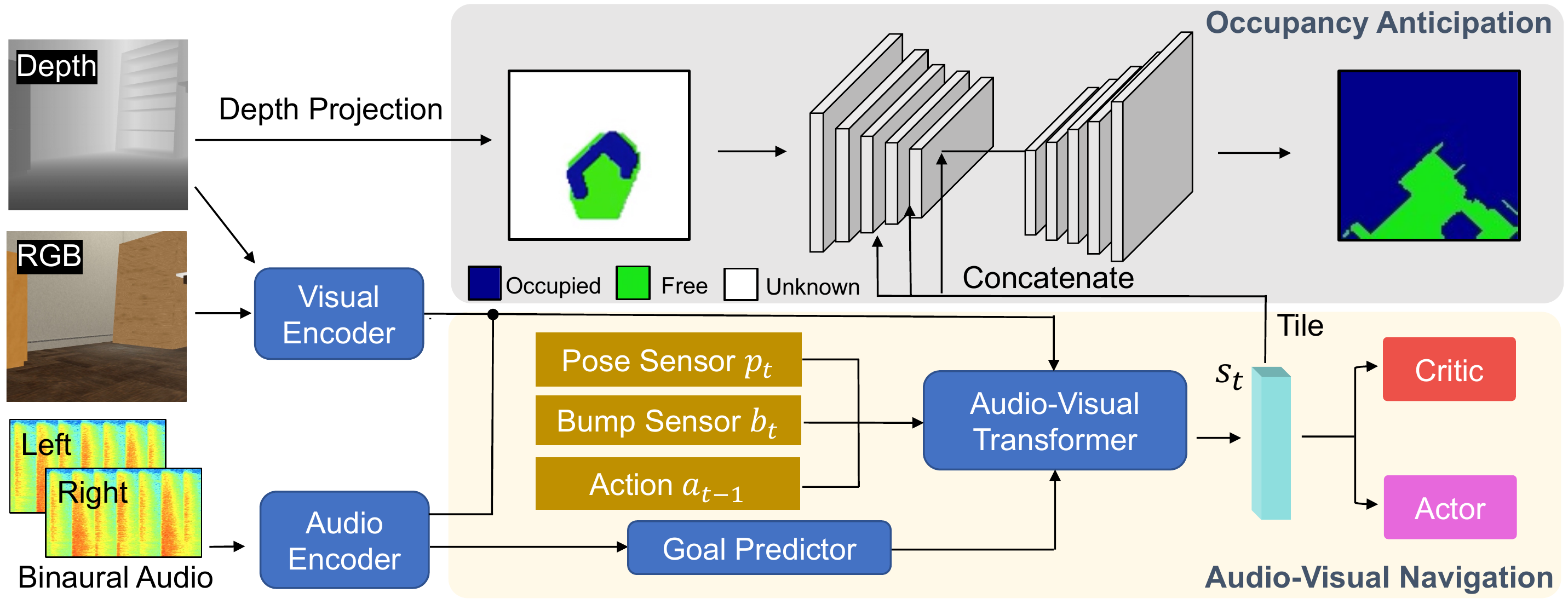}
    \caption{Our multi-task learning framework for audio-visual navigation.
    We propose to jointly learn to predict the occupancy maps and the next action to reach the goal as the agent navigates in the environment. The two tasks are mutually beneficial. The better audio-visual state features $s_t$ learned for navigation, the more useful they are for occupancy prediction; the prediction of occupancy in turn regularizes the training for audio-visual navigation.\vspace{-0.1in}}
    \vspace{-0.1in}
    \label{fig:model}
\end{figure*}

\paragraph{Audio-Visual Navigation Model.} We propose a multi-task learning framework to jointly learn for semantic audio-visual navigation and occupancy map prediction, as shown in Fig.~\ref{fig:model}. At each time step $t$, the agent receives egocentric visual observation consisting of an RGB image and a depth map, and binaural audio at the agent's left and right ears represented as a two-channel audio spectrogram. We extract the visual and audio features from the visual encoder and the audio encoder, respectively.

For semantic audio-visual navigation, we adopt the base architecture from SAVi~\cite{chen2021semantic}, adapted from the scene memory transformer network~\cite{fang2019scene}. It mainly consists of two components: 1) \emph{Goal Predictor}, which takes the audio feature and the agent's current pose as input to predict a goal descriptor that contains information about the sound source location and the object category of the sound; and 2) \emph{Audio-Visual Transformer}, which uses a memory module to encode the agent's observations and uses a self-attention mechanism to reason about the 3D environment seen so far. The decoder of the transformer takes the output of the goal predictor and the encoded observations in its memory, and predicts the state feature $s_t$, which is then fed to an actor-critic network for predicting the next action $a_t$. We follow the two-stage training paradigm in~\cite{chen2021semantic} using decentralized distributed proximal policy optimization~\cite{wijmans2019dd}. 

For occupancy map prediction, we formulate it as a pixel-wise classification task, following~\cite{ramakrishnan2020occupancy}. We represent the egocentric occupancy as a top-down map $p \in [0, 1]^{V \times V}$, which comprises a local area of $V \times V$ cells in front of the camera that represents a region of 5m $\times$ 5m. The value in each cell represents the probability of the cell being occupied. The ground-truth local occupancy is obtained by using the 3D meshes of the corresponding indoor environments. We use a U-Net~\cite{ronneberger2015u} for occupancy map prediction. The input to the encoder is the local occupancy map obtained from depth projection by setting height thresholds on the point cloud obtained from depth and camera intrinsics~\cite{chen2019learning}. We then replicate and tile the state feature vector to match the spatial dimension of the feature maps, and concatenate along the channel dimension for the last three layers of the encoder. The decoder then takes the fused feature map as input and outputs the predicted local occupancy map through a series of up-convolutional layers for both the visible and invisible cells. We use the binary cross-entropy loss for training the occupancy anticipation network.

Our occupancy map anticipation module is similar to prior methods in robotics and embodied visual navigation that build continuous representations of the world~\cite{o2012gaussian,ramos2016hilbert,katyal2019uncertainty,shrestha2019learned,ramakrishnan2020occupancy}. However, we jointly learn occupancy anticipation and audio-visual navigation, with the new insight that predicting accurate occupancy maps helps to learn better audio-visual features useful for navigation.
\section{Experiments}\label{sec:results}
We show our experiment results on audio-visual navigation,  and how we transfer agents trained in our \name simulator to real-world environments.
\begin{table*}[t]
  \centering
  {\resizebox{0.85\linewidth}{!}{
    \begin{tabular}{l|ccc|ccc|ccc|ccc}
    \toprule
    &   \multicolumn{6}{ c| }{\textit{iGibson}} & \multicolumn{6}{ c }{\textit{Matterport3D}} \\
    &   \multicolumn{3}{ c| }{\textit{Heard}} &  \multicolumn{3}{ c| }{\textit{Unheard}} & \multicolumn{3}{ c| }{\textit{Heard}} &  \multicolumn{3}{ c }{\textit{Unheard}} \\
    Model  & {SR } & {SPL} & {SNA} & {SR} & {SPL } & {SNA} & {SR } & {SPL} & {SNA} & {SR } & {SPL } & {SNA}\\
    \midrule
    Random Agent                         &  2.4  &  2.4  &  1.2  & 2.4  & 2.4  & 1.2 & 1.5  &  1.5 & 1.2 &  1.5  & 1.5  & 1.2  \\
    Gan et al.~\cite{gan2019look}                         &  12.1  &  9.2  &  8.9  & 5.0  & 3.9  & 2.9 & 5.9   &  3.8 & 3.5 &  4.1  & 3.5  & 2.7 \\
    Chen et al.~\cite{chen2020soundspaces}             &  41.2  &  39.0  &  9.7  & 24.2  & 22.0  & 4.8 & 18.1   &  17.1 & 5.2 &  10.0  & 9.4  & 2.3 \\
    SAVi~\cite{chen2021semantic}                          &  53.0  &  47.4  &  9.9  & 43.5  & 37.9 & 9.3 & 27.9   &  26.8 & 7.32 &  22.4  & 20.6  & 4.7 \\
     Ours         &  \textbf{60.2}  &  \textbf{53.6}  &  \textbf{13.8}  & \textbf{47.1}  & \textbf{41.9}  & \textbf{10.6} & \textbf{38.4}   & \textbf{37.7} & \textbf{10.5} &  \textbf{32.4}  & \textbf{29.4}  & \textbf{7.4} \\
    \bottomrule
  \end{tabular}
  }}
  \vspace{-0.1in}
   \caption{Semantic audio-visual navigation results on the iGibson dataset and the Matterport3D dataset. Our proposed multi-task learning framework compares favorably against the closest competitor SAVi, the current state-of-the-art method. It outperforms all other baselines that do not consider the semantic meaning of the objects by a large margin. 
   }
  \label{table:nav_results}
\end{table*}

\paragraph{Baselines.}
We evaluate on audio-visual separation and compare to a series of baseline methods~\cite{chen2020soundspaces,chen2021waypoints,gan2019look,chen2021semantic}:
\begin{itemize}
    \item Random Agent: A baseline that randomly samples an action at each time step and automatically stops after the agent reaches its goal.
    \item Gan et al.~\cite{gan2019look}: A map-based approach that predicts the goal location from audio and then has the agent navigate to the predicted location with an analytical path planner.
    \item Chen et al.~\cite{chen2020soundspaces}: An end-to-end approach for training RL navigation agents that leverage audio-visual observations, and it uses a GRU RNN to encode past memory.
    \item SAVi~\cite{chen2021semantic}:  A state-of-the-art semantic audio-visual navigation model that uses a goal descriptor network to provide both location and object category information to the agent, and uses a transformer-based policy network.
\end{itemize}

\begin{figure*}
    \center
    \includegraphics[scale=0.54]{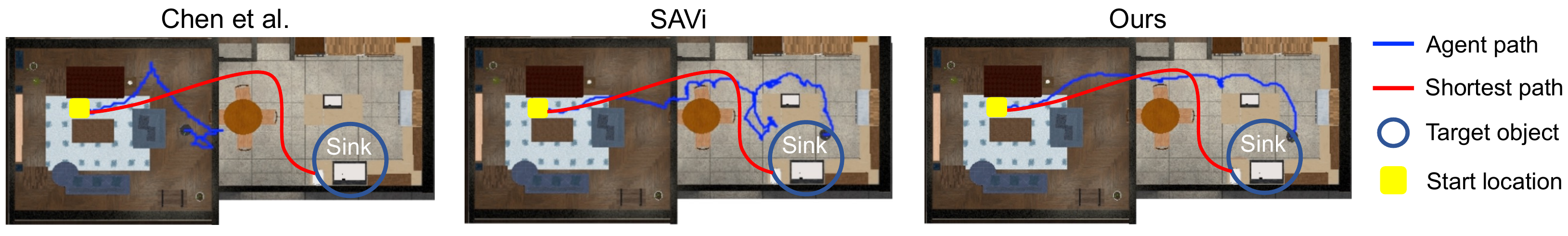}
    \caption{Navigation trajectories on top-down maps for semantic audio-visual navigation. Chen et al.~\cite{chen2020soundspaces}: The agent gets lost near the entrance of the dining room after the sound stops.  SAVi~\cite{chen2021semantic}: The agent struggles to avoid the obstacles and takes a long path to reach the goal. Ours: The agent successfully avoids all obstacles and efficiently navigates to the target object. 
    \vspace{-0.1in}}
    \vspace{-0.1in}
    \label{fig:navigation_qual}
\end{figure*}

\paragraph{Metrics.} We evaluate using the following metrics~\cite{chen2020soundspaces,chen2021semantic}: 1) success rate (SR), which is the fraction of successful episodes; 2) success weighted by (normalized inverse) Path Length (SPL)~\cite{anderson2018evaluation}, which is success times the ratio of the agent's path length to the shortest path; 3) success weighted by the inverse of the number of actions (SNA), which penalizes collisions and in-place rotations. Results for all metrics are obtained by averaging over 1,000 test trails.

\paragraph{Semantic Audio-Visual Navigation.} We use the iGibson dataset and the Matterport3D dataset for evaluation. On both datasets, we evaluate under two settings following the evaluation protocols from~\cite{chen2020soundspaces,chen2021semantic}: 1) \emph{heard} sounds---train and test on the same sound in unseen environments, and 2) \emph{unheard} sounds---train and test on disjoint sounds in unseen environments. Table~\ref{table:nav_results} shows the results. Our model significantly outperforms prior audio-visual navigation methods that do not take the semantic meaning of objects into consideration: Gan et al.~\cite{gan2019look} and Chen et al.~\cite{chen2020soundspaces}. This shows that our model successfully learns to match object categories with their sounds and leverage the semantic cues to navigate to the goal more efficiently. Compared to SAVi~\cite{chen2021semantic}---the state-of-the-art model for semantic audio-visual navigation, our multi-task learning framework consistently outperforms it across all metrics on both heard and unheard sounds, demonstrating the benefit of jointly learning to predict occupancy maps during navigation. 

Fig.~\ref{fig:navigation_qual} shows the navigation trajectories on top-down maps in a challenging scene of our model and two baseline methods. The agent of Chen et al.~\cite{chen2020soundspaces} does not consider the semantic meaning of the sound source, and thus gets lost near the entrance of the dining room after the sound stops. The SAVi agent can successfully navigate to the goal, but it takes a much longer path compared to our method. Our agent can better sense the obstacles and the sound that travels through the door---potentially due to better audio-visual state features learned jointly with occupancy anticipation---and efficiently navigate to the target object (sink). 

\paragraph{Sim2Real Transfer.} To demonstrate the realism of our \name simulator, we transfer the audio-visual navigation agents trained in simulation to real-world environments. To ease Sim2Real transfer, we train an AudioGoal navigation agent that only uses depth and audio for navigation, and an AudioPointGoal navigation agent that additionally takes as input a GPS pointer towards the target~\cite{chen2020soundspaces}. We use ROS's built-in GMapping SLAM algorithm to obtain the TurtleBot's pose as a replacement of the GPS used during training.

During our real-world deployment, we find three steps essential to reduce the Sim2Real gap for successful policy transfer. First, since the TurtleBot makes ego-noise during movement, we record the sounds of the robot moving in various environments, and mix the source sound with a random segment of the noise recording during training. Second, we also randomly vary the source gain as additional data augmentation due to the mismatch of the sound volume in simulation and real-world experiments. Third, we calibrate the depth camera to the range of 0.8m-3.5m to be consistent with the setting in simulation.

Table~\ref{tab:sim2real} shows the Sim2Real results in the bedroom shown in Fig.~\ref{fig:Sim2Real}. We perform 10 test trials with different robot starting locations or sound source locations. We can see that both models transfer reasonably well to the real world, and AudioPointGoal has a higher success rate due to the additional target pointer that complements the audio signal. To our best knowledge, this is the first result that shows audio-visual navigation policies trained in simulation can be successfully transferred to real-world environments, demonstrating the realism of our multisensory simulation. See Supp. video for qualitative examples.

\begin{table}[t]
\RawFloats
    \vspace{0.05in}
\centering
\begin{tabular}{lccc} 
        \toprule
          & SR & SPL & SNA
        \\\midrule
        AudioGoal
        & 30.0
        & 22.2
        & 5.5
        \\
        AudioPointGoal
        & 50.0
        & 31.5
        & 24.6
        \\\bottomrule
\end{tabular}
\caption{Sim2Real results for audio-visual navigation.}
\label{tab:sim2real}
\vspace{-0.15in}
\end{table}

\section{Conclusion}
We presented \name, a multisensory simulation platform for training household agents that can both see and hear. Our simulator can render continuous audio in 3D environments in real-time and also support audio streaming with VR, providing a new testbed for many embodied AI tasks that need audio-visual perception. Using the audio-visual navigation task as a case study, we propose a new model for semantic audio-visual navigation, outperforming a series of prior methods. Furthermore, we successfully deploy agents trained in simulation into real-world environments. We look forward to the embodied multisensory learning research that will be enabled by \name.




\small
\section*{ACKNOWLEDGMENT}
This work is in part supported by ONR MURI N00014-22-1-2740, ONR MURI N00014-21-1-2801, NSF \#2120095, Stanford Institute for Human-Centered AI (HAI), Adobe, Amazon, Bosch, Meta, and Salesforce.


\bibliographystyle{ieeetran}
\bibliography{ref}

\begin{thebibliography}{10}
\providecommand{\url}[1]{#1}
\csname url@rmstyle\endcsname
\providecommand{\newblock}{\relax}
\providecommand{\bibinfo}[2]{#2}
\providecommand\BIBentrySTDinterwordspacing{\spaceskip=0pt\relax}
\providecommand\BIBentryALTinterwordstretchfactor{4}
\providecommand\BIBentryALTinterwordspacing{\spaceskip=\fontdimen2\font plus
\BIBentryALTinterwordstretchfactor\fontdimen3\font minus
  \fontdimen4\font\relax}
\providecommand\BIBforeignlanguage[2]{{%
\expandafter\ifx\csname l@#1\endcsname\relax
\typeout{** WARNING: IEEEtran.bst: No hyphenation pattern has been}%
\typeout{** loaded for the language `#1'. Using the pattern for}%
\typeout{** the default language instead.}%
\else
\language=\csname l@#1\endcsname
\fi
#2}}

\bibitem{gupta2017cognitive}
S.~Gupta, J.~Davidson, S.~Levine, R.~Sukthankar, and J.~Malik, ``Cognitive
  mapping and planning for visual navigation,'' in \emph{CVPR}, 2017.

\bibitem{habitat19iccv}
M.~Savva, A.~Kadian, O.~Maksymets, Y.~Zhao, E.~Wijmans, B.~Jain, J.~Straub,
  J.~Liu, V.~Koltun, J.~Malik, D.~Parikh, and D.~Batra, ``Habitat: {A}
  {P}latform for {E}mbodied {AI} {R}esearch,'' in \emph{ICCV}, 2019.

\bibitem{wijmans2019dd}
E.~Wijmans, A.~Kadian, A.~Morcos, S.~Lee, I.~Essa, D.~Parikh, M.~Savva, and
  D.~Batra, ``Dd-ppo: Learning near-perfect pointgoal navigators from 2.5
  billion frames,'' in \emph{ICLR}, 2020.

\bibitem{Chaplot2020Learning}
D.~S. Chaplot, D.~Gandhi, S.~Gupta, A.~Gupta, and R.~Salakhutdinov, ``Learning
  to explore using active neural slam,'' in \emph{ICLR}, 2020.

\bibitem{zhu-iccv2017}
Y.~Zhu, D.~Gordon, E.~Kolve, D.~Fox, L.~Fei-Fei, A.~Gupta, R.~Mottaghi, and
  A.~Farhadi, ``{Visual Semantic Planning using Deep Successor
  Representations},'' in \emph{ICCV}, 2017.

\bibitem{chaplot_object_2020}
D.~S. Chaplot, D.~Gandhi, A.~Gupta, and R.~Salakhutdinov, ``Object goal
  navigation using goal-oriented semantic exploration,'' in \emph{NeurIPS},
  2020.

\bibitem{savinov2018semi}
N.~Savinov, A.~Dosovitskiy, and V.~Koltun, ``Semi-parametric topological memory
  for navigation,'' in \emph{ICLR}, 2018.

\bibitem{zhou2019towards}
X.~Zhou, Y.~Gao, and L.~Guan, ``Towards goal-directed navigation through
  combining learning based global and local planners,'' \emph{Sensors}, 2019.

\bibitem{Xiang_2020_SAPIEN}
F.~Xiang, Y.~Qin, K.~Mo, Y.~Xia, H.~Zhu, F.~Liu, M.~Liu, H.~Jiang, Y.~Yuan,
  H.~Wang, L.~Yi, A.~X. Chang, L.~J. Guibas, and H.~Su, ``{SAPIEN}: A simulated
  part-based interactive environment,'' in \emph{CVPR}, 2020.

\bibitem{shen2020igibson}
B.~Shen, F.~Xia, C.~Li, R.~Mart{\'\i}n-Mart{\'\i}n, L.~Fan, G.~Wang, S.~Buch,
  C.~D'Arpino, S.~Srivastava, L.~P. Tchapmi, \emph{et~al.}, ``igibson, a
  simulation environment for interactive tasks in large realisticscenes,'' in
  \emph{IROS}, 2021.

\bibitem{robothor}
M.~Deitke, W.~Han, A.~Herrasti, A.~Kembhavi, E.~Kolve, R.~Mottaghi,
  J.~Salvador, D.~Schwenk, E.~VanderBilt, M.~Wallingford, L.~Weihs, M.~Yatskar,
  and A.~Farhadi, ``{RoboTHOR: An Open Simulation-to-Real Embodied AI
  Platform},'' in \emph{CVPR}, 2020.

\bibitem{puig2018virtualhome}
X.~Puig, K.~Ra, M.~Boben, J.~Li, T.~Wang, S.~Fidler, and A.~Torralba,
  ``Virtualhome: Simulating household activities via programs,'' in
  \emph{CVPR}, 2018.

\bibitem{gan2020threedworld}
C.~Gan, J.~Schwartz, S.~Alter, M.~Schrimpf, J.~Traer, J.~De~Freitas,
  J.~Kubilius, A.~Bhandwaldar, N.~Haber, M.~Sano, \emph{et~al.}, ``Threedworld:
  A platform for interactive multi-modal physical simulation,'' in
  \emph{NeurIPS Datasets and Benchmarks Track}, 2021.

\bibitem{zhu2020robosuite}
Y.~Zhu, J.~Wong, A.~Mandlekar, and R.~Mart{\'\i}n-Mart{\'\i}n, ``robosuite: A
  modular simulation framework and benchmark for robot learning,'' \emph{arXiv
  preprint arXiv:2009.12293}, 2020.

\bibitem{james2020rlbench}
S.~James, Z.~Ma, D.~R. Arrojo, and A.~J. Davison, ``Rlbench: The robot learning
  benchmark \& learning environment,'' \emph{RA-L}, 2020.

\bibitem{chen2020soundspaces}
C.~Chen, U.~Jain, C.~Schissler, S.~V.~A. Gari, Z.~Al-Halah, V.~K. Ithapu,
  P.~Robinson, and K.~Grauman, ``Sound{S}paces: Audio-visual navigation in 3d
  environments,'' in \emph{ECCV}, 2020.

\bibitem{li2021igibson}
C.~Li, F.~Xia, R.~Mart{\'\i}n-Mart{\'\i}n, M.~Lingelbach, S.~Srivastava,
  B.~Shen, K.~Vainio, C.~Gokmen, G.~Dharan, T.~Jain, \emph{et~al.}, ``igibson
  2.0: Object-centric simulation for robot learning of everyday household
  tasks,'' in \emph{CoRL}, 2021.

\bibitem{gan2019look}
C.~Gan, Y.~Zhang, J.~Wu, B.~Gong, and J.~B. Tenenbaum, ``Look, listen, and act:
  Towards audio-visual embodied navigation,'' in \emph{ICRA}, 2020.

\bibitem{chen2021waypoints}
C.~Chen, S.~Majumder, A.-H. Ziad, R.~Gao, S.~Kumar~Ramakrishnan, and
  K.~Grauman, ``Learning to set waypoints for audio-visual navigation,'' in
  \emph{ICLR}, 2021.

\bibitem{chen2021semantic}
C.~Chen, Z.~Al-Halah, and K.~Grauman, ``Semantic audio-visual navigation,'' in
  \emph{CVPR}, 2021.

\bibitem{purushwalkam2020audio}
S.~Purushwalkam, S.~V.~A. Gari, V.~K. Ithapu, C.~Schissler, P.~Robinson,
  A.~Gupta, and K.~Grauman, ``Audio-visual floorplan reconstruction,'' in
  \emph{ICCV}, 2021.

\bibitem{gao2020visualechoes}
R.~Gao, C.~Chen, Z.~Al-Halab, C.~Schissler, and K.~Grauman, ``Visualechoes:
  Spatial image representation learning through echolocation,'' in \emph{ECCV},
  2020.

\bibitem{szot2021habitat}
A.~Szot, A.~Clegg, E.~Undersander, E.~Wijmans, Y.~Zhao, J.~Turner, N.~Maestre,
  M.~Mukadam, D.~Chaplot, O.~Maksymets, A.~Gokaslan, V.~Vondrus, S.~Dharur,
  F.~Meier, W.~Galuba, A.~Chang, Z.~Kira, V.~Koltun, J.~Malik, M.~Savva, and
  D.~Batra, ``Habitat 2.0: Training home assistants to rearrange their
  habitat,'' in \emph{NeurIPS}, 2021.

\bibitem{kolve2017ai2}
E.~Kolve, R.~Mottaghi, W.~Han, E.~VanderBilt, L.~Weihs, A.~Herrasti, D.~Gordon,
  Y.~Zhu, A.~Gupta, and A.~Farhadi, ``Ai2-thor: An interactive 3d environment
  for visual ai,'' \emph{arXiv preprint arXiv:1712.05474}, 2017.

\bibitem{yu2020meta}
T.~Yu, D.~Quillen, Z.~He, R.~Julian, K.~Hausman, C.~Finn, and S.~Levine,
  ``Meta-world: A benchmark and evaluation for multi-task and meta
  reinforcement learning,'' in \emph{CoRL}, 2020.

\bibitem{chen22soundspaces2}
C.~Chen, C.~Schissler, S.~Garg, P.~Kobernik, A.~Clegg, P.~Calamia, D.~Batra,
  P.~W. Robinson, and K.~Grauman, ``Soundspaces 2.0: A simulation platform for
  visual-acoustic learning,'' \emph{arXiv}, 2022.

\bibitem{batra2020objectnav}
D.~Batra, A.~Gokaslan, A.~Kembhavi, O.~Maksymets, R.~Mottaghi, M.~Savva,
  A.~Toshev, and E.~Wijmans, ``Objectnav revisited: On evaluation of embodied
  agents navigating to objects,'' \emph{arXiv preprint arXiv:2006.13171}, 2020.

\bibitem{zhao2021evaluation}
M.~Zhao, P.~Anderson, V.~Jain, S.~Wang, A.~Ku, J.~Baldridge, and E.~Ie, ``On
  the evaluation of vision-and-language navigation instructions,'' \emph{arXiv
  preprint arXiv:2101.10504}, 2021.

\bibitem{anderson2018vision}
P.~Anderson, Q.~Wu, D.~Teney, J.~Bruce, M.~Johnson, N.~S{\"u}nderhauf, I.~Reid,
  S.~Gould, and A.~Van Den~Hengel, ``Vision-and-language navigation:
  Interpreting visually-grounded navigation instructions in real
  environments,'' in \emph{CVPR}, 2018.

\bibitem{chen2021topological}
K.~Chen, J.~K. Chen, J.~Chuang, M.~V{\'a}zquez, and S.~Savarese, ``Topological
  planning with transformers for vision-and-language navigation,'' in
  \emph{CVPR}, 2021.

\bibitem{hochreiter1997long}
S.~Hochreiter and J.~Schmidhuber, ``Long short-term memory,'' \emph{Neural
  computation}, vol.~9, no.~8, pp. 1735--1780, 1997.

\bibitem{dean2020see}
V.~Dean, S.~Tulsiani, and A.~Gupta, ``See, hear, explore: Curiosity via
  audio-visual association,'' in \emph{NeurIPS}, 2020.

\bibitem{gan2020noisy}
C.~Gan, X.~Chen, P.~Isola, A.~Torralba, and J.~B. Tenenbaum, ``Noisy agents:
  Self-supervised exploration by predicting auditory events,'' \emph{arXiv
  preprint arXiv:2007.13729}, 2020.

\bibitem{gao2021ObjectFolder}
R.~Gao, Y.-Y. Chang, S.~Mall, L.~Fei-Fei, and J.~Wu, ``Objectfolder: A dataset
  of objects with implicit visual, auditory, and tactile representations,'' in
  \emph{CoRL}, 2021.

\bibitem{gao2022ObjectFolderV2}
R.~Gao, Z.~Si, Y.-Y. Chang, S.~Clarke, J.~Bohg, L.~Fei-Fei, W.~Yuan, and J.~Wu,
  ``Objectfolder 2.0: A multisensory object dataset for sim2real transfer,'' in
  \emph{CVPR}, 2022.

\bibitem{owens2016ambient}
A.~Owens, J.~Wu, J.~H. McDermott, W.~T. Freeman, and A.~Torralba, ``Ambient
  sound provides supervision for visual learning,'' in \emph{ECCV}, 2016.

\bibitem{arandjelovic2017look}
R.~Arandjelovic and A.~Zisserman, ``Look, listen and learn,'' in \emph{ICCV},
  2017.

\bibitem{owens2018audio}
A.~Owens and A.~A. Efros, ``Audio-visual scene analysis with self-supervised
  multisensory features,'' in \emph{ECCV}, 2018.

\bibitem{Korbar2018cotraining}
B.~Korbar, D.~Tran, and L.~Torresani, ``Co-training of audio and video
  representations from self-supervised temporal synchronization,'' in
  \emph{NeurIPS}, 2018.

\bibitem{gao2018objectSounds}
R.~Gao, R.~Feris, and K.~Grauman, ``Learning to separate object sounds by
  watching unlabeled video,'' in \emph{ECCV}, 2018.

\bibitem{zhao2018sound}
H.~Zhao, C.~Gan, A.~Rouditchenko, C.~Vondrick, J.~McDermott, and A.~Torralba,
  ``The sound of pixels,'' in \emph{ECCV}, 2018.

\bibitem{gan2020music}
C.~Gan, D.~Huang, H.~Zhao, J.~B. Tenenbaum, and A.~Torralba, ``Music gesture
  for visual sound separation,'' in \emph{CVPR}, 2020.

\bibitem{gao2019coseparation}
R.~Gao and K.~Grauman, ``Co-separating sounds of visual objects,'' in
  \emph{ICCV}, 2019.

\bibitem{tzinis2020into}
E.~Tzinis, S.~Wisdom, A.~Jansen, S.~Hershey, T.~Remez, D.~P. Ellis, and J.~R.
  Hershey, ``Into the wild with audioscope: Unsupervised audio-visual
  separation of on-screen sounds,'' in \emph{ICLR}, 2021.

\bibitem{Senocak_2018_CVPR}
A.~Senocak, T.-H. Oh, J.~Kim, M.-H. Yang, and I.~So~Kweon, ``Learning to
  localize sound source in visual scenes,'' in \emph{CVPR}, 2018.

\bibitem{tian2018audio}
Y.~Tian, J.~Shi, B.~Li, Z.~Duan, and C.~Xu, ``Audio-visual event localization
  in unconstrained videos,'' in \emph{ECCV}, 2018.

\bibitem{hu2020discriminative}
D.~Hu, R.~Qian, M.~Jiang, X.~Tan, S.~Wen, E.~Ding, W.~Lin, and D.~Dou,
  ``Discriminative sounding objects localization via self-supervised
  audiovisual matching,'' in \emph{NeurIPS}, 2020.

\bibitem{gao2019visualsound}
R.~Gao and K.~Grauman, ``2.5d visual sound,'' in \emph{CVPR}, 2019.

\bibitem{morgadoNIPS18}
P.~Morgado, N.~Vasconcelos, T.~Langlois, and O.~Wang, ``Self-supervised
  generation of spatial audio for 360${}^\circ$ video,'' in \emph{NeurIPS},
  2018.

\bibitem{garg2021geometry}
R.~Garg, R.~Gao, and K.~Grauman, ``Geometry-aware multi-task learning for
  binaural audio generation from video,'' in \emph{BMVC}, 2021.

\bibitem{gorzel2019efficient}
M.~Gorzel, A.~Allen, I.~Kelly, J.~Kammerl, A.~Gungormusler, H.~Yeh, and
  F.~Boland, ``Efficient encoding and decoding of binaural sound with resonance
  audio,'' in \emph{AES International Conference on Immersive and Interactive
  Audio}, 2019.

\bibitem{Matterport3D}
A.~Chang, A.~Dai, T.~Funkhouser, M.~Halber, M.~Niessner, M.~Savva, S.~Song,
  A.~Zeng, and Y.~Zhang, ``Matterport3d: Learning from rgb-d data in indoor
  environments,'' \emph{3DV}, 2017.

\bibitem{OpenVR}
ValveSoftware, ``Openvr,'' \url{https://github.com/ValveSoftware/openvr}.

\bibitem{fang2019scene}
K.~Fang, A.~Toshev, L.~Fei-Fei, and S.~Savarese, ``Scene memory transformer for
  embodied agents in long-horizon tasks,'' in \emph{CVPR}, 2019.

\bibitem{ramakrishnan2020occupancy}
S.~K. Ramakrishnan, Z.~Al-Halah, and K.~Grauman, ``Occupancy anticipation for
  efficient exploration and navigation,'' in \emph{ECCV}, 2020.

\bibitem{ronneberger2015u}
O.~Ronneberger, P.~Fischer, and T.~Brox, ``U-net: Convolutional networks for
  biomedical image segmentation,'' in \emph{International Conference on Medical
  image computing and computer-assisted intervention}, 2015.

\bibitem{chen2019learning}
T.~Chen, S.~Gupta, and A.~Gupta, ``Learning exploration policies for
  navigation,'' \emph{arXiv preprint arXiv:1903.01959}, 2019.

\bibitem{o2012gaussian}
S.~T. O’Callaghan and F.~T. Ramos, ``Gaussian process occupancy maps,''
  \emph{The International Journal of Robotics Research}, 2012.

\bibitem{ramos2016hilbert}
F.~Ramos and L.~Ott, ``Hilbert maps: Scalable continuous occupancy mapping with
  stochastic gradient descent,'' \emph{The International Journal of Robotics
  Research}, 2016.

\bibitem{katyal2019uncertainty}
K.~Katyal, K.~Popek, C.~Paxton, P.~Burlina, and G.~D. Hager,
  ``Uncertainty-aware occupancy map prediction using generative networks for
  robot navigation,'' in \emph{ICRA}, 2019.

\bibitem{shrestha2019learned}
R.~Shrestha, F.-P. Tian, W.~Feng, P.~Tan, and R.~Vaughan, ``Learned map
  prediction for enhanced mobile robot exploration,'' in \emph{ICRA}, 2019.

\bibitem{anderson2018evaluation}
P.~Anderson, A.~Chang, D.~S. Chaplot, A.~Dosovitskiy, S.~Gupta, V.~Koltun,
  J.~Kosecka, J.~Malik, R.~Mottaghi, M.~Savva, \emph{et~al.}, ``On evaluation
  of embodied navigation agents,'' \emph{arXiv preprint arXiv:1807.06757},
  2018.

\end{thebibliography}

\end{document}